\documentclass{isprs} % isprs class modified 23-04-2019 (Dennis Wittich)
\usepackage{subfigure}
\usepackage{setspace}
\usepackage{geometry} % added 27-02-2014 Markus Englich
\usepackage{epstopdf}
\usepackage{epstopdf}
\usepackage{amsmath}
\usepackage{graphicx}
\usepackage[labelsep=period]{caption}  % added 14-04-2016 Markus Englich - Recommendation by Sebastian Brocks
\usepackage[british]{babel} 
\usepackage[hang]{footmisc}
\usepackage{dblfloatfix}
\usepackage[colorlinks=true, linkcolor=blue, citecolor=blue, urlcolor=blue, breaklinks=true]{hyperref}
\usepackage{booktabs}

\begin{document}

\title{
Point-level Uncertainty Evaluation of Mobile Laser Scanning Point Clouds
}

% KAO: Remove extra spacing
\author{Ziyang Xu\textsuperscript{1}, Olaf Wysocki\textsuperscript{3}, Christoph Holst\textsuperscript{1,2} }

% KAO: Remove extra newline
\address{%
\centering
\parbox{\textwidth}{\centering
\textsuperscript{1} Chair of Engineering Geodesy, TUM School of Engineering and Design, Technical University of Munich, 80333 Munich, Germany — \{ziyang.xu, christoph.holst\}@tum.de\\[2pt]
\textsuperscript{2} TUM Leonhard Obermeyer Center, Technical University of Munich, 80333 Munich, Germany\\
\textsuperscript{3} CV4DT, University of Cambridge, Cambridgeshire CB3 0FA, Cambridge, UK — {okw24@cam.ac.uk}
}}

\abstract{

Reliable quantification of uncertainty in Mobile Laser Scanning (MLS) point clouds is essential for ensuring the accuracy and credibility of downstream applications such as 3D mapping, modeling, and change analysis. Traditional backward uncertainty modeling heavily rely on high-precision reference data, which are often costly or infeasible to obtain at large scales. To address this issue, this study proposes a machine learning-based framework for point-level uncertainty evaluation that learns the relationship between local geometric features and point-level errors. The framework is implemented using two ensemble learning models, Random Forest (RF) and XGBoost, which are trained and validated on a spatially partitioned real-world dataset to avoid data leakage. Experimental results demonstrate that both models can effectively capture the nonlinear relationships between geometric characteristics and uncertainty, achieving mean ROC-AUC values above 0.87. The analysis further reveals that geometric features describing elevation variation, point density, and local structural complexity play a dominant role in predicting uncertainty. The proposed framework offers a data-driven perspective of uncertainty evaluation, providing a scalable and adaptable foundation for future quality control and error analysis of large-scale point clouds.

}

\maketitle
%\saythanks % added 28-02-2014 Markus Englich

\section{Introduction}\label{Introduction}
 
% KAO: Sloppy spacing ensures non-overfull lines. Can be removed if this is not an issue.
\sloppy

Mobile Laser Scanning (MLS) systems have become indispensable in fields such as digital twinning, smart cities, engineering geodesy, 3D modeling, and autonomous driving, owing to their efficiency in data collection and wide applicability \cite{Xue2020,xu2025c}. Yet, despite this progress, the point clouds acquired by MLS systems operating in real-world environments inevitably contain uncertainty arising from various error sources during acquisition and processing. Although MLS systems have advanced rapidly in both data collection and post-processing, research on uncertainty evaluation has received comparatively less attention and remains underdeveloped \cite{xu2025}.

From a user’s perspective, the quality of point clouds from MLS systems is a critical concern. As the foundational input for many downstream tasks, inadequate assessment of MLS point clouds' quality can easily impact high-precision applications such as navigation and change analysis. This will not only undermine reliability but also result in substantial waste of time and resources, which is unacceptable in real-world applications. There is a clear need for automated and reliable solutions for uncertainty evaluation. 

In MLS systems, four main categories of error sources contribute to uncertainty: instrumental errors, atmospheric errors, object- and geometry-related errors, and trajectory estimation errors \cite{Habib2009,Schenk2001}. Considering the characteristics of these error sources, existing uncertainty evaluation methods can be broadly divided into two categories: forward modeling and backward modeling \cite{Shi2021}. The core idea of forward modeling is grounded in variance-covariance propagation, which involves detailed theoretical analysis of MLS system errors. This approach models various error sources during data acquisition to infer the error characteristics of the final point clouds. However, the uncertainty of MLS point clouds is also influenced by many hardly predictable factors, including the observation environment, surface materials, internal data processing algorithms, platform speed, and scanning trajectory. As a result, accurate and comprehensive modeling of all error sources in MLS data acquisition is nearly impossible \cite{Holst2016,Shi2021}.

In contrast to forward modeling, backward modeling, as a more popular solution, bypasses complex error modeling and propagation during data collection and processing. Instead, it employs accurate reference data to empirically evaluate the errors in MLS point clouds. Many works have been done based on the cloud-to-cloud (C2C) distance, the cloud-to-mesh (C2M) distance as well as the multiscale model-to-model cloud comparison (M3C2) distance \cite{Lague2013,Heinz2015,Stenz2017,Zahs2022,Xu2025b}. To address the dependency on point cloud density and enhance the automation of the process inherent in point-based methods, line-based methods have been adopted \cite{Poreba2013,Hauser2016}. Many researchers also use plane features and have initiated several studies focused on uncertainty evaluation based on plane correspondences \cite{Shi2021,Shahraji2020,xu2025}. In summary, the key advantage of the backward strategy is that uncertainty is assessed based on actual measurements, providing a foundation for developing generalized evaluation solutions applicable across different systems. The core of backward modeling lies in establishing reliable correspondences between scanned and reference data and characterizing their inconsistencies.

It is worth noting that although backward modeling offers clear advantages and practical feasibility over forward modeling, its inherent limitations are also evident. These limitations primarily appear in two aspects:

\begin{enumerate}
\setlength\itemsep{0em}\setlength\parskip{0em}\setlength\topsep{0em}\setlength\partopsep{0em}\setlength\parsep{0em} 
\item{\textbf{Heavily depends on reference data.} For areas without reference data, backward modeling methods are ineffective because they heavily rely on the availability of high-precision reference data.}

\item{\textbf{High cost to get reference data.} In many cases, the cost of acquiring such a high-precision reference far exceeds that of MLS data collection and is often infeasible to acquire.}
\end{enumerate}

These limitations collectively restrict the scalability and general applicability. In this context, machine learning offers a promising alternative, as it enables the establishment of relationships between point cloud features and their associated uncertainty metrics \cite{hartmann2023a,hartmann2023b}. By shifting from explicit error modeling or reference-based evaluation to data-driven prediction, machine learning opens a pathway toward more flexible, efficient, and broadly applicable uncertainty evaluation solutions for MLS point clouds. This paper primarily discusses how to perform point-level uncertainty evaluation based on machine learning in the absence of reference data. It should be clarified that reference data is utilized during the training stage but is no longer required once training is complete. The main contributions are as follows:

%\itemize
\begin{itemize}
\setlength\itemsep{0em}\setlength\parskip{0em}\setlength\topsep{0em}\setlength\partopsep{0em}\setlength\parsep{0em} 
\item{A machine learning-based framework is introduced to achieve reliable point-level uncertainty prediction.} 

\item{A comprehensive experiment on a typical indoor dataset that shows the accuracy of the proposed framework.} 

\item{An investigation on feature importance in point-level uncertainty evaluation, thereby providing new insights into error source identification.}
\end{itemize}

\section{Methodology}\label{Methodology}

\begin{figure*}[!hbt]
\centering
\includegraphics[width=0.8\textwidth]{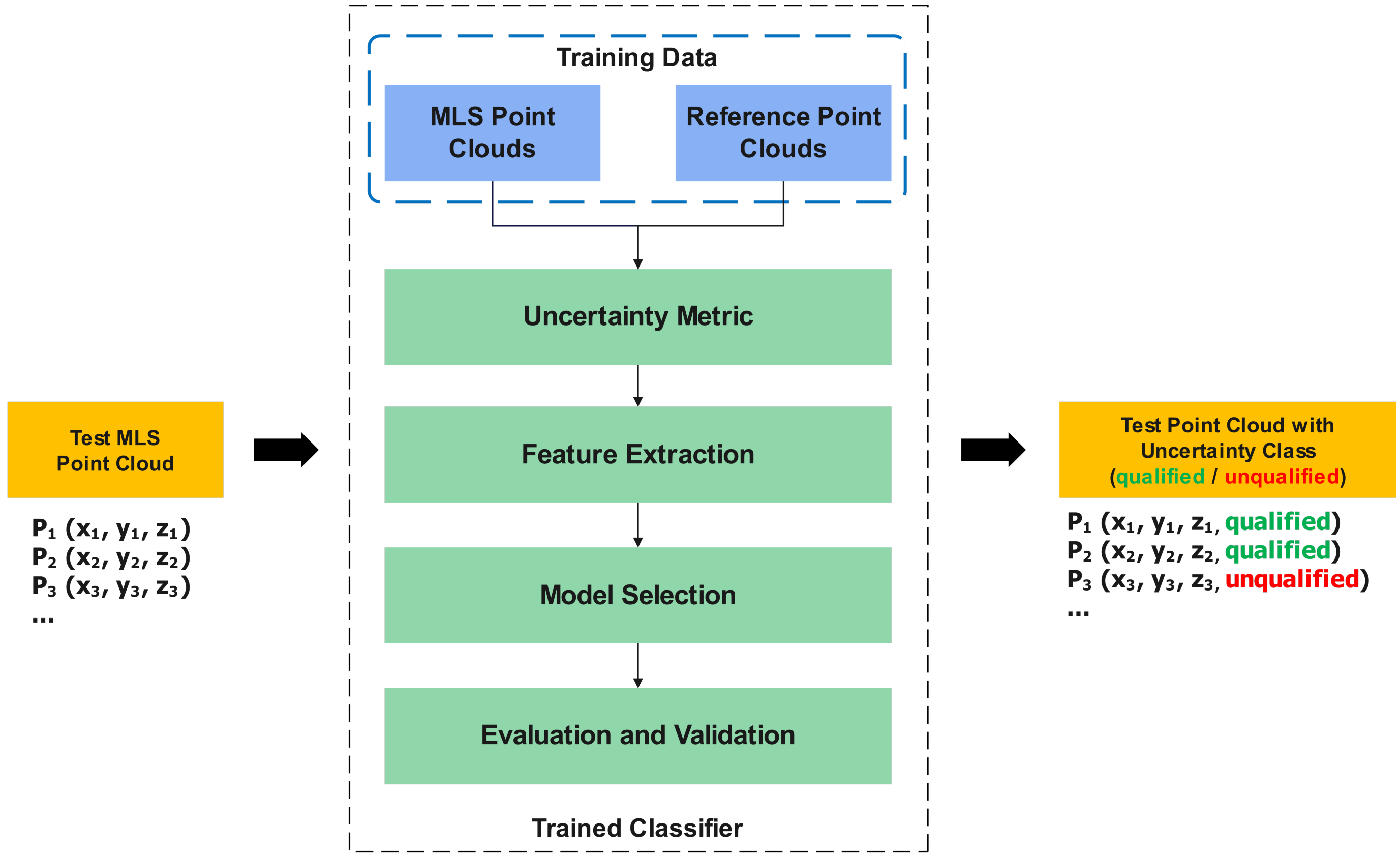}
\caption{Workflow of the proposed machine learning-based uncertainty evaluation framework.}
\label{fig:workflow}
\end{figure*}

This section presents detailed implementation of the proposed framework, comprising four steps: uncertainty metric definition in Section \ref{Uncertainty Metric Definition}, feature extraction in Section \ref{Feature Extraction}, model selection in Section \ref{Model Selection}, evaluation and validation in Section \ref{Evaluation and Validation}. Figure \ref{fig:workflow} illustrates the conceptual workflow.

\subsection{Uncertainty Metric Definition}\label{Uncertainty Metric Definition}

In this study, point-level uncertainty is quantified using C2C distance. The decision to employ C2C distance rather than M3C2 distance for point-level uncertainty quantification is grounded in two key considerations. First, C2C distance is more sensitive to variations introduced by individual point noise, thereby offering a more direct representation of local point-level error characteristics. Second, the computation of M3C2 distance relies on a predefined radius, and different radius values can influence the results. This limitation does not occur with C2C distance.

\subsection{Feature Extraction}\label{Feature Extraction}

The objective of this study is to investigate whether the uncertainty of each point, quantified by the C2C distance, is related to its local geometric features. In other words, it examines whether the error behavior of a point can be predicted or estimated from its geometric characteristics. To this end, point-level feature extraction serves as a critical step.

Furthermore, geometric feature extraction requires an appropriately defined neighborhood. Only with a well-specified neighborhood,
local geometric features can be reliably computed at the point level. However, determining the optimal neighborhood definition remains a critical and fundamental challenge, as different definitions can affect subsequent feature calculations and introduce additional instability into the overall workflow.

It has been demonstrated that the neighborhood definition based on the $k$-nearest neighbors (KNN) offers distinct advantages, and the geometric features derived from it can enhance the performance of point cloud classification tasks \cite{weinmann2015}. Additionally, geometric features have also been used for 3D learning task to achieve better performance \cite{tan2023,robert2023,wysocki2023}. 

Inspired by these findings, this study adopts the same optimal neighborhood estimation strategy to compute the corresponding geometric features. Specifically, for each point in the input MLS point clouds, the optimal number of neighboring points $OptN$ is first estimated. Subsequently, based on $OptN$, the corresponding geometric features are calculated. The detailed feature definition and computational procedure fall outside the scope of this paper; readers are referred to the relevant literature for more information \cite{weinmann2014,weinmann2015}.

\subsection{Model Selection}\label{Model Selection}

Following the extraction of local geometric features, the next step is to establish a predictive relationship between these features and the point-level quality attributes. Instead of directly regressing the uncertainty magnitude (C2C distance), this study formulates the task as a binary classification problem: determining whether each point meets the quality threshold (qualified) or not (unqualified). 

To enable classification, each point was assigned a binary label according to its uncertainty level. Specifically, the uncertainty of each point was quantified by its C2C distance with respect to the reference data. Points with a C2C distance smaller than a given threshold $t_{c}$ were labeled as \textit{qualified}, indicating acceptable quality, while those exceeding $t_{c}$ were labeled as \textit{unqualified}. This $t_{c}$ was empirically determined based on the accuracy requirements of the dataset, usecase, and the inherent noise characteristics of the MLS system. Such a criterion converts the continuous uncertainty values into discrete quality categories, allowing the problem to be effectively modeled as a binary classification task.

The input features are explicit, structured, and of moderate dimensionality, making this formulation well-suited for supervised learning. Considering the nonlinear and potentially heterogeneous relationships between geometric characteristics and point quality, ensemble tree-based classifiers are particularly appropriate. Therefore, Random Forest (RF) \cite{breiman2001} and XGBoost \cite{chen2016} were adopted to evaluate the feasibility and robustness of point-level quality classification based on local geometric features.

More specifically, RF was adopted as a stable baseline due to its robustness to noise and outliers, low sensitivity to parameter tuning, and strong resistance to overfitting. By aggregating multiple decision trees through a bagging mechanism, RF effectively reduces variance while maintaining interpretability. This property is particularly desirable for point clouds with heterogeneous local geometries, where spatial variability may otherwise degrade model performance. 

On the other hand, XGBoost is a gradient-boosted framework that sequentially builds trees to minimize residual errors using gradient descent optimization. It incorporates regularization terms to prevent overfitting and typically achieves higher predictive accuracy on structured tabular data. XGBoost has been proven effective at utilizing point cloud attributes such as intensity and incidence angle to predict Terrestrial Laser Scanning (TLS) ranging errors \cite{hartmann2023a,hartmann2023b}. Furthermore, XGBoost provides native tools for model interpretability, such as feature importance scores and SHAP value analysis, which are essential for understanding the influence of individual geometric feature on uncertainty estimation.

In summary, RF and XGBoost were chosen for their complementary advantages: RF offers a robust and interpretable reference model, while XGBoost provides stronger fitting capacity. Employing both models enables a balanced and comprehensive assessment of predictive performance and feature relevance in the context of point-level uncertainty estimation.

\subsection{Evaluation and Validation}\label{Evaluation and Validation}

The performance of two models was quantitatively evaluated using five complementary metrics: area under the receiver operating characteristic curve (ROC-AUC), average precision (AP), Precision, Recall, and F1-score. Among them, ROC-AUC and AP were computed based on the predicted class probabilities, while Precision, Recall, and F1-score were derived from the discrete classification results.

The ROC-AUC metric assesses the overall discriminative capability of the classifier across all possible threshold settings. It is defined as the area under the receiver operating characteristic (ROC) curve, which plots the true positive rate (TPR) against the false positive rate (FPR):

\begin{equation}
\mathrm{ROC\text{-}AUC} = \int_{0}^{1} \mathrm{TPR}(x)\, d(\mathrm{FPR}(x)),
\end{equation}
where $\mathrm{TPR} = \frac{TP}{TP + FN}$ and $\mathrm{FPR} = \frac{FP}{FP + TN}$.

The average precision (AP) summarizes the precision-recall (PR) curve as the weighted mean of precisions achieved at different recall levels:
\begin{equation}
\mathrm{AP} = \sum_{n} (R_{n} - R_{n-1}) \, P_{n},
\end{equation}
where $P_{n}$ and $R_{n}$ denote the precision and recall at the $n$-th threshold, respectively.

Precision and recall were computed from the binary predictions as:
\begin{equation}
\mathrm{Precision} = \frac{TP}{TP + FP}, \qquad
\mathrm{Recall} = \frac{TP}{TP + FN}.
\end{equation}
Precision reflects the reliability of positive predictions, whereas recall measures the robustness.

The F1-score provides a harmonic mean of Precision and Recall, offering a balanced measure of classification performance when the two classes are imbalanced:
\begin{equation}
\mathrm{F1} = 2 \times \frac{\mathrm{Precision} \times \mathrm{Recall}}{\mathrm{Precision} + \mathrm{Recall}}.
\end{equation}

Moreover, a 5-fold grid-based strategy was employed. Instead of performing random sampling, the entire point cloud was first divided into non-overlapping spatial grids. These grids were then grouped into five spatially distinct folds. In each iteration, four folds were used for training and the remaining one for testing. This design combines the robustness of cross-validation with the spatial independence of grid partitioning, effectively preventing data leakage while maintaining statistical reliability. The averaged results across the five folds were reported to provide a comprehensive and spatially unbiased evaluation of model performance.

\section{Experiment}\label{Experiment}

This section presents the experimental setup and analysis for the proposed framework. It includes three main components: data preparation in Section \ref{Data Preparation}, model training in Section \ref{Model Training}, and results analysis in Section \ref{Results Analysis}.

\subsection{Data Preparation}\label{Data Preparation}

\begin{figure*}[ht!]
\begin{center}
		\includegraphics[width=0.7\textwidth]{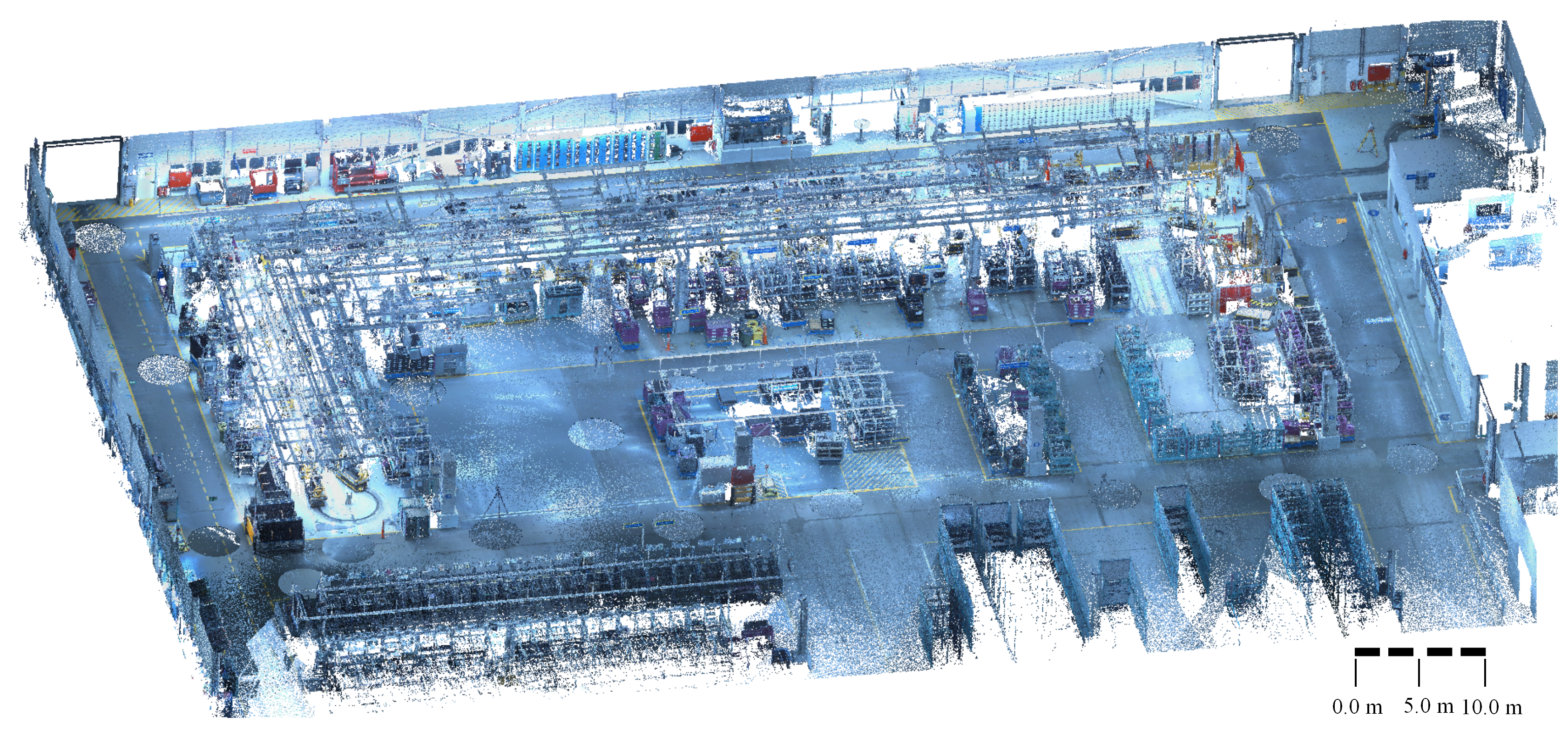}
	\caption{Overview of the scanned area.}
\label{fig:overview}
\end{center}
\end{figure*}

The experiment was conducted in an assembly hall of the BMW factory, representing a typical indoor industrial environment characterized by a combination of planar and complex structures, as well as a variety of reflective and absorbent materials. Numerous machines, shelves, boxes, and other components contributed to intricate geometries and partially occluded regions. The point clouds were acquired using a Smart Transport Robot (STR) platform equipped with a MLS system, specifically the Emesent Hovermap ST-X \footnote{\url{https://emesent.com/hovermap-series/}}. For reference data, the Trimble X9 3D laser scanning system \footnote{\url{ https://geospatial.trimble.com/en/products/hardware/trimble-x9}} was employed to obtain a high-precision reference by minimizing registration errors as much as possible. Additionally, a high-precision total station control network comprising 12 black \& white targets was established to ensure precise spatial alignment between the TLS and MLS point clouds. More information about the experiment scenario can be found in Table \ref{experiment scenario}.  An overview of the scanned area is shown in Figure \ref{fig:overview}.

\begin{table}[h]
\centering
\resizebox{\linewidth}{!}{%
\begin{tabular}{ll}
\toprule
  Scanning Area & About 3,500 m$^{2}$ \\  % 将 m^2 放入数学模式
  Number of Points & 5,000,000 \\
  MLS System & EMESENT HOVERMAP ST-X\ \\
  Accuracy Specification* & $\pm$15mm \\
 \bottomrule
  {\small * manufacturer's specification in typical environment.}\\
  \end{tabular}
}
\caption{Basic information on experiment scenario}
\label{experiment scenario}
\end{table}

After data collection and preprocessing, the C2C distance of each point in the MLS point clouds was computed using the precisely aligned TLS reference point clouds. For practical relevance, only MLS points with C2C distances below 100\,mm were retained for subsequent analysis. To enable binary classification, a distance threshold $t_{d}$ was then defined to distinguish between the \textit{qualified} and \textit{unqualified} classes. In this study, a threshold $t_{d}$ of 20\,mm was adopted, meaning that all MLS points with C2C distances smaller than 20\,mm were labeled as \textit{qualified}, while the remaining points were labeled as \textit{unqualified}. This $t_{d}$ was determined empirically, taking into account the accuracy specifications and typical error characteristics of the MLS system under the given experimental conditions.

\subsection{Model Training}\label{Model Training}

As mentioned before, to ensure spatial independence, RF and XGBoost models were both trained under the five-fold scheme to ensure comparability. For each fold, the spatially independent training and testing subsets were standardized using z-score normalization.

To prevent excessive memory usage when handling large-scale point clouds with RF, a random subsampling procedure was applied whenever the number of training samples exceeded one million, retaining approximately 30\% of the data for model fitting. The main hyperparameters used in the RF experiments are summarized in Table \ref{tab:rf_params}. The averaged results across all folds were reported to provide a spatially unbiased assessment of classification performance.

\begin{table}[h!]
\centering
\small
\resizebox{\linewidth}{!}{%
\begin{tabular}{ll}
\toprule
\textbf{Parameter} & \textbf{Value / Description} \\
\midrule
\texttt{n\_estimators} & 100 (number of trees in the forest) \\
\texttt{max\_depth} & 20 (maximum depth of each tree) \\
\texttt{max\_samples} & 0.5 (fraction of samples used per tree) \\
\texttt{class\_weight} & \texttt{balanced} (adjusts weights inversely to class frequency) \\
\texttt{n\_jobs} & 1 (single-threaded training for memory stability) \\
\texttt{random\_state} & fixed random seed for reproducibility \\
\bottomrule
\end{tabular}
}
\caption{Main hyperparameters used for Random Forest.}
\label{tab:rf_params}
\end{table}

The XGBoost model was trained on a NVIDIA GeForce RTX 3070 Laptop GPU (8 GB) using a binary logistic objective function and evaluated with the log-loss metric. Model training was performed with up to 1000 boosting rounds, with early stopping after 50 rounds if the validation loss did not improve. The main hyperparameters used in the experiments are summarized in Table \ref{tab:xgb_params}.

\begin{table}[h!]
\centering
\small
\resizebox{\linewidth}{!}{%
\begin{tabular}{ll}
\toprule
\textbf{Parameter} & \textbf{Value / Description} \\
\midrule
\texttt{objective} & \texttt{binary:logistic} (binary classification) \\
\texttt{eval\_metric} & \texttt{logloss} \\
\texttt{max\_depth} & 8 (maximum tree depth) \\
\texttt{eta} (learning rate) & 0.05 \\
\texttt{subsample} & 0.8 (row subsampling ratio) \\
\texttt{colsample\_bytree} & 0.8 (feature subsampling ratio per tree) \\
\texttt{num\_boost\_round} & 1000 (maximum boosting iterations) \\
\texttt{early\_stopping\_rounds} & 50 (validation-based early stopping) \\
\texttt{tree\_method} & \texttt{hist} (GPU-accelerated histogram algorithm) \\
\texttt{scale\_pos\_weight} & adaptive negative-to-positive ratio \\
\texttt{seed} & fixed random seed \\
\bottomrule
\end{tabular}
}
\caption{Main hyperparameters used for XGBoost.}
\label{tab:xgb_params}
\end{table}

\subsection{Results Analysis}\label{Results Analysis}

\begin{figure*}[h!]
  \centering
  \includegraphics[width=0.8\textwidth]{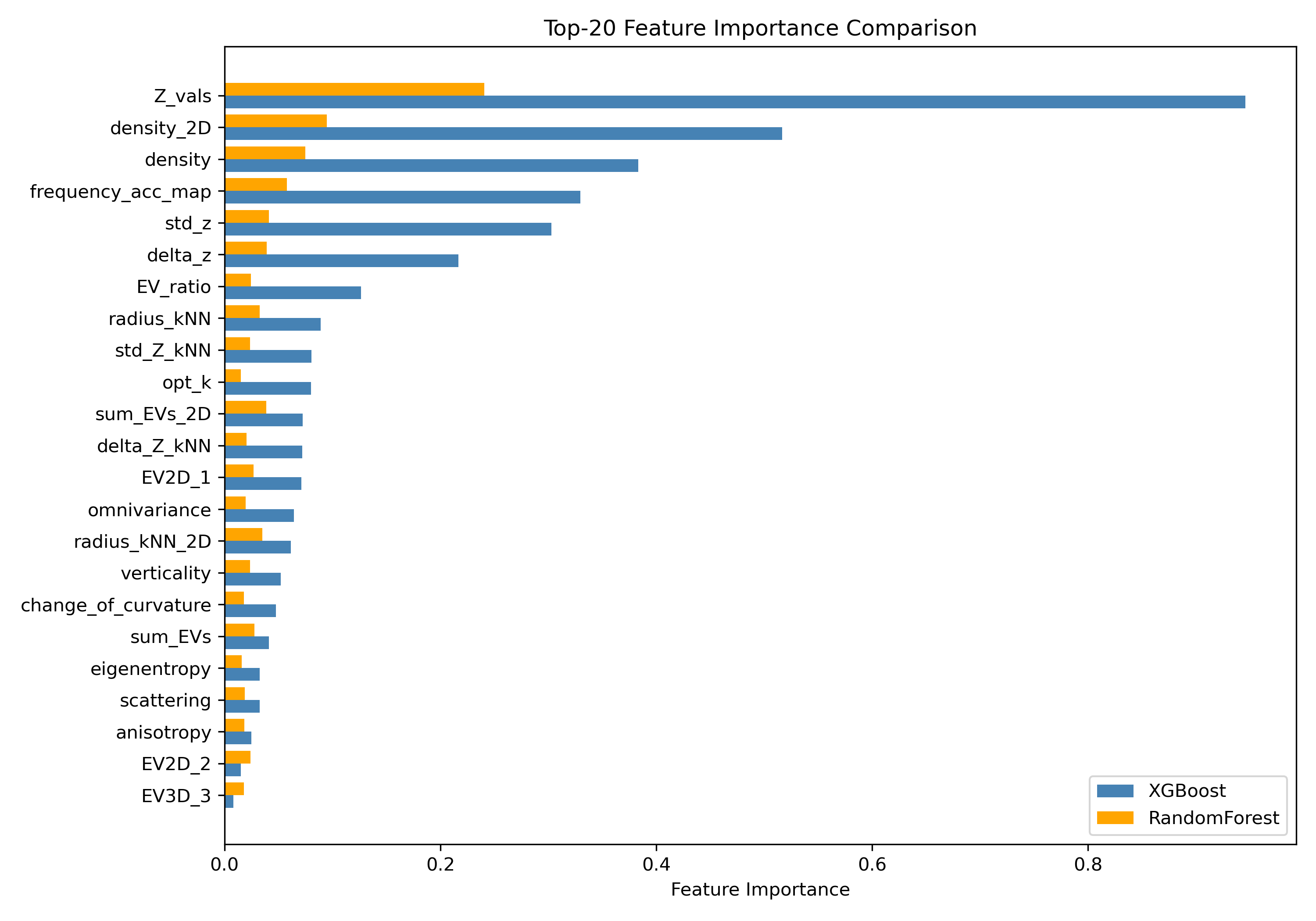}
  \caption{Comparison of the top 20 most important features.}
  \label{fig:feature_importance_top20}
\end{figure*}

Table \ref{tab:rf_results} summarizes the averaged classification results across all folds. Although the RF classifier achieved relatively high ROC-AUC (0.8745) and AP (0.7158) values, its Precision, Recall, and F1-scores at the default threshold of 0.5 were comparatively lower (around 0.63). This indicates that the model possesses strong overall discriminative capability. It is able to rank points by their likelihood of belonging to the \textit{qualified} class. But the fixed decision threshold does not optimally separate the two categories. One reason behind this might be that the geometric characteristics of points near the 2\,cm boundary are highly similar, leading to overlapping feature distributions and increased misclassifications. Consequently, while the model’s probabilistic predictions are reliable, its hard classification performance is limited by the continuous nature of the uncertainty distribution in the input data.

\begin{table}[h!]
\centering
\small
\resizebox{\linewidth}{!}{%
\begin{tabular}{lcc}
\toprule
\textbf{Metric} & \textbf{Mean $\pm$ 95\% CI} & \textbf{Fold values} \\
\midrule
ROC-AUC & $0.8745 \pm 0.0130$ & [0.8576, 0.8719, 0.8818, 0.8775, 0.8835] \\
AP & $0.7158 \pm 0.0202$ & [0.7028, 0.7004, 0.7109, 0.7384, 0.7267] \\
Precision@0.5 & $0.6264 \pm 0.0184$ & [0.6478, 0.6107, 0.6146, 0.6265, 0.6324] \\
Recall@0.5 & $0.6419 \pm 0.0350$ & [0.6050, 0.6339, 0.6316, 0.6774, 0.6615] \\
F1@0.5 & $0.6337 \pm 0.0173$ & [0.6257, 0.6221, 0.6230, 0.6510, 0.6466] \\
\bottomrule
\end{tabular}
}
\caption{Performance of the Random Forest.}
\label{tab:rf_results}
\end{table}

The XGBoost classifier was evaluated using the same strategy to ensure direct comparability. As shown in Table \ref{tab:xgb_results}, XGBoost achieved a mean ROC-AUC of 0.8840 and an AP of 0.7407, both slightly higher than those of the RF. This indicates that the boosting-based approach provided stronger discriminative ability and better ranking consistency between qualified and unqualified samples. Although its Precision was moderate, the model maintained high Recall (0.7785) and balanced F1-scores, demonstrating effective identification of qualified points across heterogeneous spatial regions. Overall, XGBoost exhibited superior classification capability while maintaining robustness and stability across folds.

\begin{table}[h!]
\centering
\small
\resizebox{\linewidth}{!}{%
\begin{tabular}{lcc}
\toprule
\textbf{Metric} & \textbf{Mean $\pm$ 95\% CI} & \textbf{Fold values} \\
\midrule
ROC-AUC & $0.8840 \pm 0.0148$ & [0.8643, 0.8840, 0.8940, 0.8849, 0.8929] \\
AP & $0.7407 \pm 0.0209$ & [0.7252, 0.7259, 0.7377, 0.7646, 0.7503] \\
Precision@0.5 & $0.5340 \pm 0.0203$ & [0.5590, 0.5143, 0.5267, 0.5339, 0.5360] \\
Recall@0.5 & $0.7785 \pm 0.0235$ & [0.7520, 0.7690, 0.7788, 0.7941, 0.7985] \\
F1@0.5 & $0.6332 \pm 0.0134$ & [0.6413, 0.6164, 0.6284, 0.6385, 0.6414] \\
\bottomrule
\end{tabular}
}
\caption{Performance of the XGBoost.}
\label{tab:xgb_results}
\end{table}

While the preceding analysis focuses on the overall classification performance of the RF and XGBoost, it is also essential to understand the underlying factors that drive these predictions. Interpreting feature importance enables a deeper examination of how different geometric attributes contribute to the model’s decision-making process and helps to identify which features are most influential in distinguishing between qualified and unqualified points. In this regard, both RF and XGBoost offer inherent mechanisms for feature interpretation: RF derives feature importance from the average impurity reduction across trees, whereas XGBoost provides more refined gradient-based importance measures that can be further complemented by SHAP value analysis for local interpretability. 

Figure \ref{fig:feature_importance_top20} presents a side-by-side comparison of the top 20 most influential geometric features ranked by the XGBoost and RF. Both consistently identified \texttt{Z\_vals}, \texttt{density\_2D}, and \texttt{density} as the dominant predictors, followed by \texttt{frequency\_acc\_map}, \texttt{std\_z}, and \texttt{delta\_z}. These features primarily reflect local elevation variation, point distribution uniformity, and geometric variability, which strongly influence the C2C distance itself. Since the C2C metric is used to quantify point-level uncertainty, features that affect local geometric deviation naturally exhibit higher predictive importance in both models. In other words, the observed feature relevance originates from the inherent sensitivity of the C2C measurement to local surface irregularities, point spacing, and neighborhood configuration, rather than from external measurement conditions or scanner parameters.

While the overall ranking trends were highly consistent between the two models, XGBoost assigned relatively higher importance to the top-ranked features, indicating stronger concentration of predictive power within a few dominant descriptors. In contrast, the RF distributed importance values more evenly across features, suggesting enhanced robustness against feature redundancy and noise. Together, these results highlight that both models capture the same underlying geometric dependencies, yet differ slightly in how feature influence is weighted across the ensemble.

\begin{figure}[h!]
  \centering
  \includegraphics[width=1.0\linewidth]{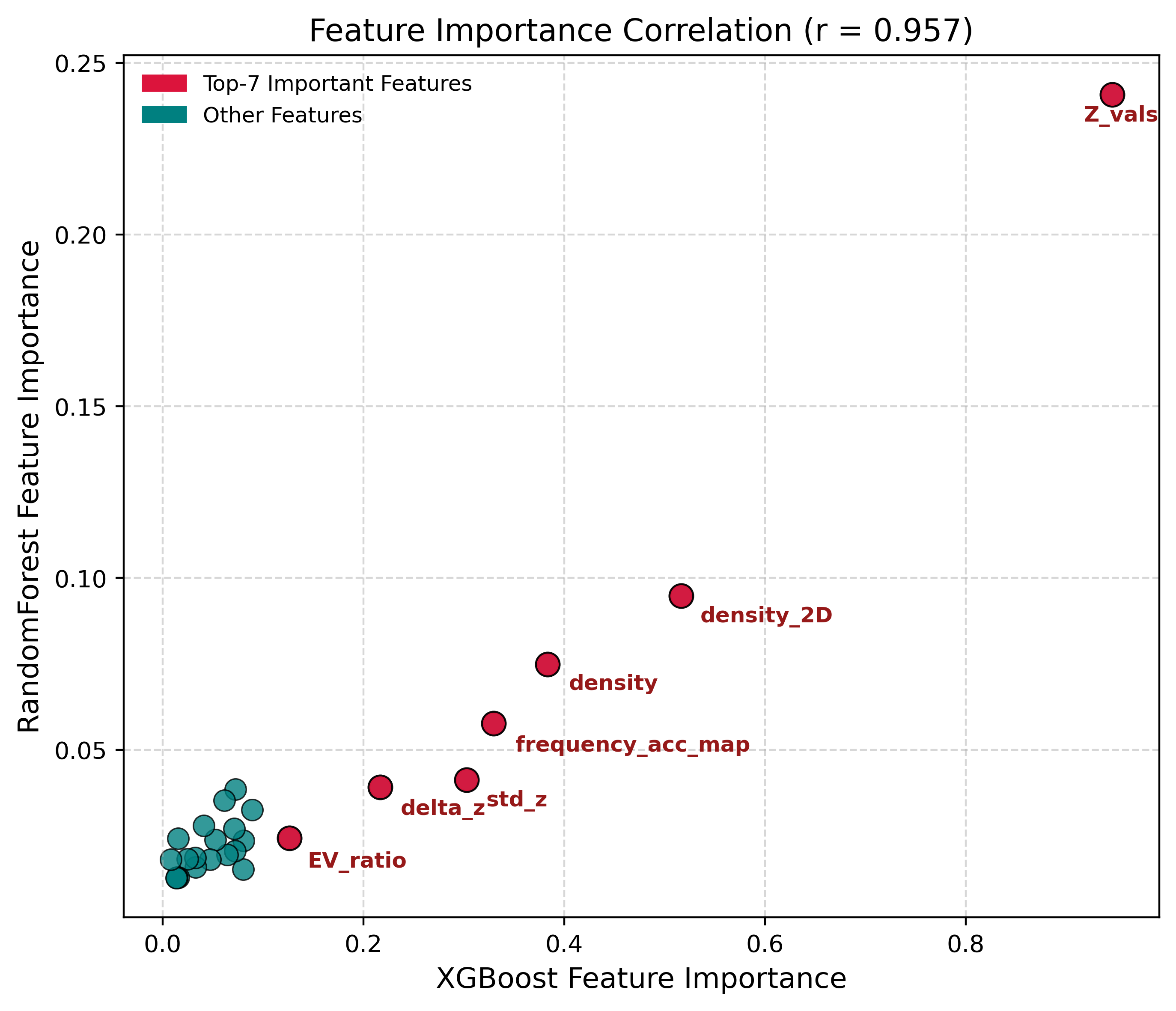}
  \caption{Correlation between feature importance scores of two models. For better visualization, the axes of RF and XGBoost are unequal.}
  \label{fig:feature_importance_corr}
\end{figure}

To further investigate the consistency between the two ensemble models, the feature importance scores derived from the RF and XGBoost were compared, as illustrated in Figure \ref{fig:feature_importance_corr}. A strong linear correlation ($r = 0.957$) was observed between their importance rankings, indicating again that both models identified largely the same geometric features as the dominant factors influencing classification results. The strong agreement between the two models confirms the robustness of the feature set and validates that the learned patterns are not model-specific but rather inherent to the geometric properties of the data or the uncertainty metrics used.

\section{Discussion}\label{Discussion}

This section presents the advantages of the proposed framework in \ref{Advantages} and challenges which should be further investigated in \ref{Challenges}.

\subsection{Advantages}\label{Advantages}
The primary advantage of this study lies in exploring and preliminarily demonstrating the feasibility of learning and predicting point-level errors directly from point cloud geometric features. A data-driven learning framework is established, marking a substantial and promising step toward learning-based uncertainty evaluation. This finding suggests that the long-standing and challenging problem of point cloud uncertainty evaluation can be approached from a feature-learning perspective, thereby gradually reducing the dependence on high-precision reference data.

Moreover, the proposed framework exhibits strong scalability and flexibility, particularly in the definition of uncertainty metrics, the selection of geometric features, and the choice of learning models. These characteristics provide a solid foundation for subsequent in-depth investigations and potential extensions.

\subsection{Challenges}\label{Challenges}

Although the proposed framework demonstrates promising performance about distinguishing whether a point is qualified or unqualified, several challenges remain that warrant further attention. First, the model’s capability in multi-class classification tasks has not yet been explored. How to enable the framework to move beyond binary classification to more fine-grained error categorization remains an open question. Second, the selective extraction of the most informative geometric features from a large pool of candidates, with the goal of improving computational efficiency and interpretability, also requires further investigation. Finally, the current study has been validated only within a single experimental setting and dataset; thus, its cross-scenario applicability and generalization ability should be systematically examined in future work.

\section{Conclusion}\label{Conclusion}

This study proposed a machine learning-based framework for point-level uncertainty evaluation in MLS point clouds, aiming to reduce the dependence on high-precision reference data in conventional backward strategy. By learning from geometric features that inherently describe local structural and spatial characteristics, the framework successfully demonstrated that point-level errors can be predicted in a data-driven manner. Two ensemble learning models, RF and XGBoost, were employed to validate the idea. Both models achieved stable and consistent performance across spatially independent folds, confirming the feasibility of predicting uncertainty directly from point-level geometric descriptors. 

The results indicate that point-level uncertainty is strongly correlated with local geometric variability and distribution uniformity. They are also factors that inherently influence the C2C distance used as the uncertainty metric. This finding provides empirical evidence that uncertainty evaluation can be reframed as a feature-learning problem, enabling the integration of data-driven methods into quality control pipelines for large-scale point cloud applications. The proposed framework is flexible and scalable, allowing adaptation to different feature definitions, uncertainty metrics, and learning algorithms. Future work will focus on extending the framework to multi-class uncertainty categorization, optimizing feature selection for computational efficiency, and validating its generalization across diverse scenarios and datasets.

\section*{CRediT authorship contribution statement}
\textbf{Ziyang Xu}: Conceptualization, Methodology, Software, Validation, Investigation, Writing – review \& editing, Writing – original draft, Visualization. \textbf{Olaf Wysocki}: Conceptualization, Writing - Review \& Editing. \textbf{Christoph Holst}: Supervision, Writing - Review \& Editing, Funding acquisition, Project administration.

\section*{Declaration of competing interest}
The authors declare that they have no known competing financial interests or personal relationships that could have appeared to influence the work reported in this paper.

\section*{Acknowledgment}
This research was funded by TUM Georg Nemetschek Institute of Artificial Intelligence for the Built World, project "NERF2BIM”, PI Christoph Holst.

{
	\begin{spacing}{1.17}
		\normalsize
		\bibliography{GeoBench} % Include your own bibliography (*.bib), style is given in isprs.cls
	\end{spacing}
}

\end{document}